%% file: main.tex
\documentclass[sigconf,authorversion,nonacm]{acmart}

\usepackage{array}
\newcolumntype{H}{>{\setbox0=\hbox\bgroup}c<{\egroup}@{}}
\usepackage[T1]{fontenc}
\usepackage[utf8]{inputenc}
\usepackage{hyperref}
\usepackage{amsmath}
\usepackage{adjustbox}
\usepackage{arydshln}
\usepackage{todonotes}
\usepackage{multirow}

\newcommand{\dovspace}{\vspace{0pt}}

\AtBeginDocument{%
  }

\setcopyright{acmcopyright}
\copyrightyear{2023}
\acmYear{2023}
\acmDOI{XXXXXXX.XXXXXXX}

\acmConference[BDA '23]{}{}{}
\acmPrice{0.00}
\acmISBN{0}

\begin{document}

\title{Mapping and Cleaning Open Commonsense Knowledge Bases with Generative Translation}

\author{Julien Romero}
\email{julien.romero@telecom-sudparis.eu}
\affiliation{%
  \institution{Télécom SudParis, SAMOVAR, IP Paris}
   \country{France}
}

\author{Simon Razniewski}
\email{Simon.Razniewski@de.bosch.com}
\affiliation{%
  \institution{Bosch Center for AI}
   \country{Germany}
}

\renewcommand{\shortauthors}{Romero \& Razniewski}

\begin{abstract}
  \input{parts/abstract}

\end{abstract}






\newcommand{\sr}[1]{\textcolor{green}{SR: #1}} 

\renewcommand{\paragraph}[1]{\smallskip\noindent\textbf{#1.\mbox{\ \ }}}

\maketitle

\input{parts/introduction}

\input{parts/previous_work}

\input{parts/problem}

\input{parts/methodology}

\input{parts/experiment_setup}

\input{parts/results}

\input{parts/conclusion}

\bibliographystyle{ACM-Reference-Format}
\bibliography{bib}

\end{document}

%% file: parts/abstract.tex
Structured knowledge bases (KBs) are the backbone of many know\-ledge-intensive applications, and their automated construction has received considerable attention. In particular, open information extraction (OpenIE) is often used to induce structure from a text. However, although it allows high recall, the extracted knowledge tends to inherit noise from the sources and the OpenIE algorithm. Besides, OpenIE tuples contain an open-ended, non-canonicalized set of relations, making the extracted knowledge's downstream exploitation harder. In this paper, we study the problem of mapping an open KB into the fixed schema of an existing KB, specifically for the case of commonsense knowledge. We propose approaching the problem by \textit{generative translation}, i.e., by training a language model to generate fixed-schema assertions from open ones. Experiments show that this approach occupies a sweet spot between traditional manual, rule-based, or classification-based canonicalization and purely generative KB construction like COMET. Moreover, it produces higher mapping accuracy than the former while avoiding the association-based noise of the latter. Code and data are available at \href{https://julienromero.fr/data/GenT}{julienromero.fr/data/GenT}.

%% file: parts/introduction.tex
\section{Introduction}
\label{sec:introduction}

\paragraph{Motivation and Problem.}
Open Information Extraction (OpenIE) automatically extracts knowledge from a text. The idea is to find explicit relationships, together with the subject and the object they link. For example, from the sentence ``In nature, fish swim freely in the ocean.'', OpenIE could extract the triple \textit{(fish, swim in, the ocean)}. Here, the text explicitly mentions the subject, the predicate, and the object. Therefore, if one uses OpenIE to construct a knowledge base (we call it an Open Knowledge Base, open KB) from a longer text, one obtains many predicates, redundant statements, and ambiguity.

OpenIE is often used for commonsense knowledge base (CSKB) construction. Previous works such as TupleKB~\cite{mishra2017domain}, Quasimodo~\cite{quasimodo,10.1145/3340531.3417416} or Ascent~\cite{nguyen2021refined,nguyen2021advanced,nguyen2021inside} use OpenIE to extract knowledge from different textual sources (textbooks, query logs, question-answering forums, search engines, or the Web), and then add additional steps to clean and normalize the obtained data. Another example is ReVerb~\cite{fader2011identifying}, which was used to get OpenIE triples from a Web crawl. The output of OpenIE typically inherits noise from sources and extraction, and the resulting KBs contain an open-ended set of predicates. This generally is not the case for knowledge bases with a predefined schema. Famous instances of this type are manually constructed, like ConceptNet~\cite{conceptnet} and ATOMIC~\cite{comet-atomic-2020}. They tend to have higher precision. Besides, they are frequently used in downstream applications such as question-answering \cite{mhgrn,yasunaga2021qagnn}, knowledge-enhanced text generation~\cite{10.1145/3512467}, image classification~\cite{DBLP:conf/aaai/WangHLLZMW20}, conversation recommender systems~\cite{DBLP:conf/kdd/ZhouZBZWY20}, or emotion detection~\cite{zhong-etal-2019-knowledge}. These applications assume there are few known predicates so that we can learn specialized parameters for each relation (a matrix or embeddings with a graph neural network). This is not the case for open KBs.

Still, many properties of open KBs, such as high recall and ease of construction, are desirable. In this paper, we study \textit{how to transform an open KB into a KB with a predefined schema}. More specifically, we study the case of commonsense knowledge, where ConceptNet is by far the most popular resource. From an open KB, we want to generate a KB with the same relation names as ConceptNet. This way, we aim to increase precision and rank the statements better while keeping high recall. Notably, as we reduce the number of relations, we obtain the chance to make the statements corroborate. For example, \textit{(fish, live in, water, freq:1)}, \textit{(fish, swim in, water, freq:1)} and \textit{(fish, breath in, water, freq:1)} can be transformed into \textit{(fish, LocatedIn, water, freq:3)}, and therefore they all help to consolidate that statement. Besides, we make new KBs available to work with many existing applications originally developed for ConceptNet.

Transforming open triples to a predefined schema raises several challenges. In the simplest case, the subject and object are conserved, and we only need to predict the correct predefined predicate. This would be a classification task. For example, \textit{(fish, live in, water)} can be mapped to \textit{(fish, LocatedAt, water)} in ConceptNet. We could proceed similarly in cases where subject and object are inverted, like mapping \textit{(ocean, contain, fish)} to \textit{(fish, LocatedAt, ocean)}, with just an order detection step. However, in many cases, the object is not expressed in the same way or only partially: \textit{(fish, live in, the ocean)} can be mapped to \textit{(fish, LocatedAt, ocean)}. In other cases, part or all of the predicate is in the object, like \textit{(fish, swim in, the ocean)} that can be mapped to \textit{(fish, CapableOf, swim in the ocean)}. Here, the initial triple could also be mapped to \textit{(fish, LocatedAt, ocean)}, showing that the mapping is not always unique. Other problems also arise, like with (near) synonyms. For example, we might want to map \textit{(fish, live in, sea)} to \textit{(fish, LocatedAt, ocean)}.

\paragraph{Approach and Contribution.}
We propose to approach the mapping of an open KB to a predefined set of relations as a translation task. We start by automatically aligning triples from the source and target KB. Then, we use these alignments to finetune a generative language model (LM) on the translation task: Given a triple from an open KB, the model produces one or several triples in the target schema. The generative nature of the LM allows it to adapt to the abovementioned problems while keeping a high faithfulness w.r.t.\ the source KB. Besides, we show that this improves the precision of the original KB and provides a better ranking for the statements while keeping a high recall.

We first introduce previous works in Section~\ref{sec:previous_work}. Then, we define our problem formally in Section~\ref{sec:problem}. In Section~\ref{sec:model}, we present our methodology with the model we use and how we construct a dataset automatically. In Section~\ref{sec:experiment_setup}, we describe the setup of our experiments. In Section~\ref{sec:experiments}, we compare the models and see the advantages of using an LM-based translation model.

Our contributions are:

\begin{enumerate}
    \item We define the problem of open KB mapping, delineating it from the more generic KB canonicalization and the more specific predicate classification.
    \item We propose a generative translation model based on pre-trained language models trained on automatically constructed training data.
    \item We experimentally verify the advantages of this method compared to traditional manual and rule-based mapping, classification, and purely generative methods like COMET.
\end{enumerate}

%% file: parts/previous_work.tex
\section{Previous Work}
\label{sec:previous_work}

\subsection{Commonsense Knowledge Bases}

\paragraph{ConceptNet} ConceptNet~\cite{conceptnet}, built since the late 1990s via crowdsourcing, is arguably today's most used commonsense knowledge base. Due to user-based construction, it has high precision. ConceptNet comprises a limited set of predefined relations and contains non-disambiguated entities and concepts. For example, we find \textit{(mouse, PartOf, computer)} and \textit{(mouse, PartOf, rodent family)}. Thus, when mapping an open KB to ConceptNet, one needs to focus mainly on the predicates and, to some extent, the modification of the subject and object.

\paragraph{Open Knowledge Base} An open knowledge base (open KB) is a collection of SPO triples \textit{(subject, predicate, object)} with no further constraints on the components. This means that they are not canonicalized. For example, the triples \textit{(The Statue of Liberty, is in, New York)} and \textit{(Statue of Liberty, located in, NYC)}, although equivalent, could be present in the same knowledge base. The subject and the object are noun phrases (NP), whereas the predicate is a relational phrase (RP). As a comparison, knowledge bases with a predefined schema like Wikidata~\cite{vrandevcic2014wikidata}, YAGO~\cite{tanon2020yago} or ConceptNet~\cite{conceptnet} come with a set of predefined predicates and/or entities for the subjects and the objects. This paper will call such a knowledge base a \textit{Closed Knowledge Base} (closed KB).

\paragraph{Construction of open KBs} The construction of open KBs relies on Open Information Extraction (OpenIE) algorithms. These algorithms take as input a text and return a set of open triples such that the subject, the predicate, and the object are explicitly mentioned in the text. There exist several systems like CoreNLP~\cite{manning2014stanford} or OpenIE6~\cite{kolluru2020openie6}.

This paper will use two open KBs: Quasimodo and Ascent++. Quasimodo~\cite{quasimodo} is an open commonsense knowledge base constructed automatically from query logs and question-answering forums. Ascent++ \cite{nguyen2021refined} is also an open commonsense knowledge base created from Web content. The extraction follows a classical pipeline and outputs an open KB in both cases.

\subsection{From Open KBs to Closed KBs}

\paragraph{Open Knowledge Base Canonicalization} The task of open KB canonicalization~\cite{galarraga2014canonicalizing} consists of turning an open triple \textit{(s, p, o)}, where s and o are an NP and p is an RP, into an equivalent (semantically) new triple \textit{($s_e$, $p_e$, $o_e$)}, where $s_e$ and $o_e$ represent entities (generally through a non-ambiguous NP), and $p_e$ is a non-ambiguous and unique representation of a predicate. It means there is no other $p_e'$ such that \textit{($s_e$, $p_e$, $o_e$)} is semantically equivalent to \textit{($s_e$, $p_e'$, $o_e$)}. For example, we would like to map \textit{(Statue of Liberty, located in, NYC)} to \textit{(The Statue of Liberty, AtLocation, New York City)}, where ``The Statue of Liberty'' represents only the famous monument in New York City, ``New York City'' represents the American city unambiguously, and ``AtLocation'' is a predicate used to give the location of the subject.  

NP canonicalization is more studied than RP canonicalization, but the task is generally treated as a clusterization problem~\cite{galarraga2014canonicalizing}. It is essential to notice that an NP or an RP does not necessarily belong to a single cluster, as this cluster may depend on the context. For example, in \textit{(Obama, be, president of the US)}, ``Obama'' refers to the entity ``Barack Obama'', whereas in \textit{(Obama, wrote, Becoming)}, ``Obama'' refers to ``Michele Obama''. Also, we must notice that canonicalization does not have a target: \textit{The transformation does not try to imitate the schema of an existing knowledge base}. The main goal is to reduce redundancy, but the number of predicates (and entities) might remain high.

\paragraph{Entity Linking} Entity Linking is the task of mapping an entity name to an entity in a knowledge base. For example, we would like to map \textit{Paris} in \textit{(Paris, be, city of love)} to \href{https://www.wikidata.org/wiki/Q90}{Q90} in Wikidata, the entity that represents the capital of France. In the triple \textit{(Paris, be, a hero)}, \textit{Paris} should be mapped to \href{https://www.wikidata.org/wiki/Q167646}{Q167646} in Wikidata, the entity that represents the son of Priam. When mapping an open KB to a closed KB, most systems first perform entity linking before processing the predicate~\cite{dutta2014semantifying,zhang2019openki}. This supposes that the subject and the object remain unchanged during the mapping. This is a problem when we want to map to ConceptNet as this KB is not canonicalized, and the subject and object might be modified.

\paragraph{Knowledge Base Construction} Knowledge base construction can be done manually by asking humans to fill in the KB~\cite{conceptnet,comet-atomic-2020} or automatically using pattern matching~\cite{tanon2020yago,auer2007dbpedia} or OpenIE~\cite{quasimodo,nguyen2021refined,mishra2017domain}. In general, manual approaches have higher accuracy but struggle to scale. Translating an open KB to a closed KB can be seen as an additional stage in an OpenIE extraction pipeline like Quasimodo or Ascent++. By doing so, we make the KB match a predefined schema. The same result would be possible directly from the corpus using traditional IE techniques. However, this approach is more human-labor intensive, depends on the domain, and does the scale~\cite{zhou2022survey}.

\paragraph{Ontology Matching} Ontology matching is the task of mapping one structured schema into another \cite{euzenat2007ontology}. This task has a long history in databases and semantic web research. However, due to the input being of little variance in predicates, it is typically approached as a structured graph alignment problem \cite{doan2004ontology,euzenat2011ontology}. We cannot simply map one predicate on another in the present problem, as textual predicates are generally ambiguous. The mapping may differ for different s-o-pairs with the same p.

\input{parts/existing_systems_short}

%% file: parts/existing_systems_short.tex
\subsection{Existing Systems}

In this paper, we are interested in a task that was barely tackled by previous works: We want to map an \textit{entire} open KB to the schema of an existing closed KB. In the Ascent++ paper~\cite{nguyen2021refined}, the authors noticed that using an open KB in practice was difficult due to the lack of existing frameworks. Therefore, they proposed to map Ascent++ to ConceptNet's schema. However, they did a straightforward manual mapping that involved translating as many relations as possible manually. This approach is simplistic and does not yield good results, as we will see later. KBPearl~\cite{lin2020kbpearl} did a variation of the manual mapping in which they used the existing labels of entities and predicates, which greatly limits the system.

When we look at similar tasks, we find two main ideas to transition between an open KB and a closed KB. First, some authors approached this problem via rule mining, a generalization of the manual mapping of predicates. Previous systems~\cite{soderland2013open,soderland2010adapting,dutta2014semantifying} often use a rule mining system (automatic or manual) that relies on the type of subject and object and keywords in the triple. They often return a confidence score. The main issues with these frameworks are that they generalize poorly (particularly to unseen predicates) and require significant human work.

The second way to see our problem is as a classification task: Given an open triple \textit{(s, p, o)}, we want to predict a semantically equivalent/related triple \textit{(s, p', o)} that would be in the considered closed KB. OpenKI~\cite{zhang2019openki} used neighbor relations as input of their classifier. Later~\cite{wood2021integrating}, word embeddings were included to represent the predicate and help with the generalization. However, their training and testing dataset is constructed using an open KB and a closed KB with entities already aligned by humans (ReVerb~\cite{fader2011identifying} and Freebase in the original paper). This is not generally the case in practice. Besides, this approach considers that the subject and object remain the same, thus ignoring modification of the subject and object, inverse relations, or closely related entities.

In~\cite{putri2019aligning}, the authors propose a method to compute the similarity between a triple in an open KB and a triple in a closed KB. This differs from our approach because we do not know potential candidates in the closed KB in advance. Indeed, the closed KB is often incomplete, and we want to generate new triples thanks to the open KB. Therefore, we focus more on the generation rather than the comparison. However, it is essential to notice that this approach integrates word embeddings for comparison. Besides, the authors use the distant supervision approach to create a dataset automatically: Given a close triple, they find sentences (in a different corpus) containing both entities from the triple. Then, they apply an OpenIE algorithm to obtain an open triple. This triple is used as a ground truth. In our case, we do not have this additional textual source: The inputs are the open KB and the closed KB.

T-REx~\cite{elsahar2018t} aligns Wikipedia abstracts with the Wikidata triples using a rule-based system. However, it comes with several limitations. First, it takes as input text and not open triples. Even if we were to take the documents used for constructing Ascent++ (a web crawl), the computation time would be much longer because of the difference in scale. Second, T-REx needs to perform named-entity recognition which does not apply to commonsense. Third, there is a strong dependency between Wikipedia and Wikidata. Some pages are even created automatically from Wikidata. Despite these limitations, we can consider the rule-based alignment presented in Section~\ref{sec:alignment} as a generalization of their AllEnt aligner. T-REx was used for evaluating language models in a zero-shot fashion~\cite{wang2022deepstruct,petroni2019}, or for OpenIE~\cite{wang2021zero}.

In~\cite{gashteovski2020aligning}, the authors introduce a methodology to manually evaluate the alignment of triples from an open KB with a closed KB. Besides, they studied how much an open KB (OPIEC~\cite{gashteovski2019opiec} in their case) can be expressed by a closed KB (DBpedia~\cite{auer2007dbpedia}). They found that the open triples can often be aligned to DBpedia facts, but they are generally more specific. Also, one can usually express an OpenIE fact in the DBpedia schema. Still, this expressivity is limited if we consider only a single relation rather than a conjunction (or even a more expressive logical formula).

%% file: parts/problem.tex
\section{Problem Formulation}
\label{sec:problem}

\newcommand{\K}{\mathcal{K}}
\newcommand{\R}{\mathcal{R}}

An open triple $t$ consists of a subject $s$, a predicate $p$, and an object $o$. An open knowledge base $\K_O$ is a set of open triples. A closed schema $\R_C$ is a set of relations $\{R_1, \ldots, R_n\}$.
A triple mapping $m$ is a function that takes an open triple $t$ and a closed schema $\R_C$ and produces a set of triples with predicates from $\R_C$.

Note that $m$ is not defined as producing a single output triple per input triple - depending on the closed schema's structure, some open triples may give rise to several closed triples. Besides, the subject and object are not guaranteed to remain the same.

\paragraph{Problem} Given an open KB $\K_O$ and a closed set of relations $\R_C$, the task is to find a mapping $m$ that enables to build a closed KB $\K_C = m(\K_O,\R_C)$, with the following properties:

\begin{enumerate}
    \item \textbf{Preserves source recall}. In other words, ensure that as many triples as possible are mapped to a nonempty set, maximizing $\mid\{t_O\in\K_O \mid m(t_O,\R_C)\neq \emptyset)\mid.$
    \item \textbf{Remains source-faithful}. In other words, ensure that each triple in the output stems from one or several semantically similar statements in the input, that is, that for each $t_c\in\K_C$, $m^{-1}(t_C,\R_C)$ is semantically similar to $t_C$.
    \item \textbf{Corrects errors}. In other words, the goal is to minimize the set of triples in $\K_C$ that are factually wrong.
\end{enumerate}

The definition above hinges on the concept of semantic similarity. In line with previous work~\cite{gashteovski2020aligning}, we specifically refer to semantic equivalence or entailment: The truth of $t_O$ should be a sufficient condition for the truth of $t_C$. However, our method does not desire the opposite direction, producing $t_C$ statements that are only sufficient conditions for $t_O$.

%% file: parts/methodology.tex
\section{Methodology}
\label{sec:model}

\begin{figure}
    \centering
    \includegraphics[width=\columnwidth]{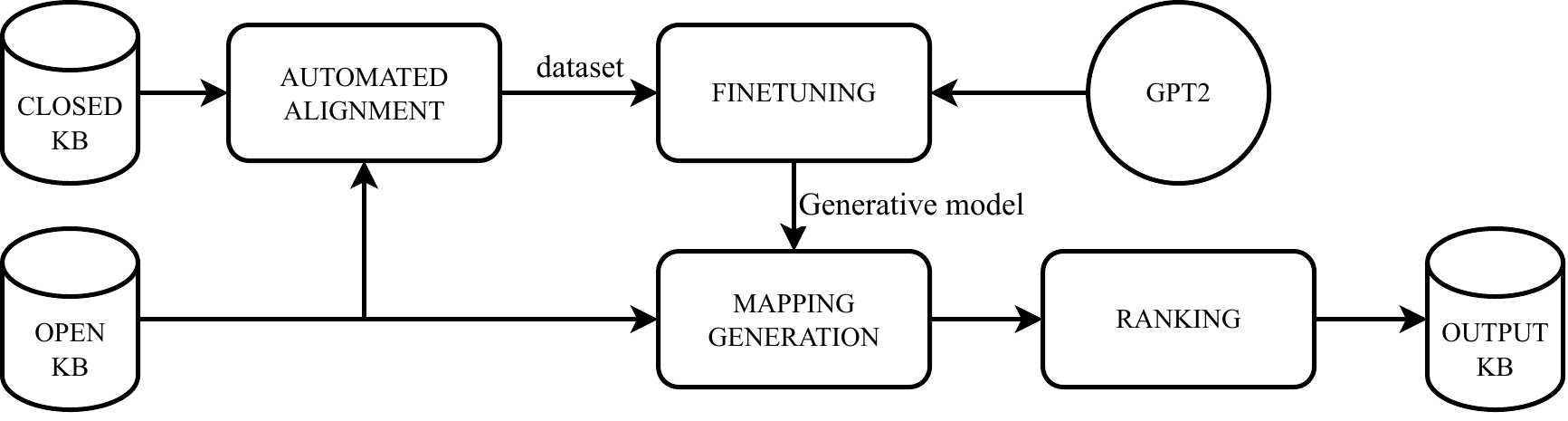}
    \caption{Our Methodology.}
    \label{fig:methodology}
\end{figure}

As we saw in Section~\ref{sec:previous_work} and will describe in more detail in Section~\ref{sec:baselines}, previous works propose to tackle the open KB mapping task in three different ways: manual mapping, rule mining mapping, or classifier mapping. However, these methods all come with challenges: They require much human work, cannot modify the subject and the object, cannot cover all cases, and, as we will see, have low performance. Therefore, we introduce here a new methodology to tackle these issues. We present in Figure~\ref{fig:methodology} our approach. It is composed of four steps:

\begin{enumerate}
    \item \textit{Alignment}: We automatically create a dataset of alignments using the open KB and the closed KB.
    \item \textit{Finetuning}: We use this dataset to finetune a generative language model (GPT-2 here) to generate alignments.
    \item \textit{Generation}: We generate one or several mappings for each triple in the open KB. 
    \item \textit{Ranking}: Using the score in the original KB, the generation score, and the rank of the generated alignment, we create a final score for each closed triple.
\end{enumerate}

This section describes in more detail how we implemented these steps.

\subsection{Creating Weakly-Labelled Training Data}
\label{sec:alignment}

The generative translation mapping and existing classification and rule mining approaches require a training dataset of alignments, that is, pairs of open triples and semantically equivalent or entailed closed triples. Creating this at scale manually is hardly feasible. Therefore, we decided to adopt two automatic approaches to generate a broader dataset, even if they contain some noise.

\subsubsection{Rule-Based Alignment}

The first approach we consider is based on rules. To align an open KB with ConceptNet, we used the following algorithm:

\begin{enumerate}
    \item Lemmatization and stopwords removal.
    \item For each triple $(s, p, o)$ in our open KB, we create one or several alignments if (we used \_ as a general placeholder):
    \begin{itemize}
        \item s, \_, o is in Conceptnet (standard alignment)
        \item o, \_, s is in Conceptnet (reverse alignement)
        \item s, \_, p + o is in Conceptnet (predicate in the object)
        \item p + o, \_, s is in Conceptnet (reverse predicate in the object)
    \end{itemize}
\end{enumerate}

This approach has the advantage of not creating divergence in the alignment: We have hard constraints (words) that do not allow us to align statements that are too different. However, this is not true with the following technique.

\subsubsection{LM-Based Alignment}

Here, we propose an entirely unsupervised method. First, we compute the embeddings of each triple in both KBs and then align each open triple with the nearest close triple. For computing the triple embeddings, we used a sentence embeddings neural network fed with the subject, the predicate, and the object, separated by a comma. The Python library SBert provides a MiniLM~\cite{wang2020minilm} model finetuned on a paraphrasing task. Then, we used Scikit-learn~\cite{pedregosa2011scikit} K-nearest neighbor algorithm to find the nearest neighbor in a closed KB for each triple in an open KB (or the opposite, marked as INV later) and the distance between the two triples. Finally, as considering all the alignments might introduce noise, we assess several scenarios in which we only take the top 1k, 10k, and 100k alignments according to the distance score. As generating the mapping is expensive, we will not finetune this parameter more.

With this technique, we might have alignments that are not related. However, compared to the previous method, we will be able to have a larger dataset, and we might get semantic relatedness coming from using different wording (e.g., with synonyms) that was not captured before.




\subsection{Generative Translation Mapping}

The second step consists in finetuning a generative language model to generate the mappings. This can be seen as a translation problem, similar to machine translation. We formatted our input by separating the OpenIE triple and its aligned ConceptNet triple with a \textit{[SEP]} token. In our experiments, we used the GPT-2 model~\cite{radford2019language} and the script provided by HuggingFace to finetune GPT-2. Unfortunately, GPT-3~\cite{brown2020language} is not publicly available. We also tried T5~\cite{raffel2020exploring}, but we did not obtain better results (see T5-GenT in Table~\ref{tab:auto_rec_prec}). We also accessed a very large language model, LLaMa~\cite{llama}, following Alpaca~\cite{alpaca,llama-adapter}. However, this model failed to adapt to the structure of the closed KB, even when given various explicit prompts. We hypothesize that such a model lost flexibility as it better understood natural language. Succeeding in reintroducing structured information in very large LM can lead to exciting future works. Another disadvantage of very large LMs is their computation cost at training and during inference.

The third step is the actual generation. We used a beam search for the generation part to obtain the top $K$ results for each statement in our knowledge base. We filter the results to keep only well-formed triples and triples so that the subject and object differ. Considering more than one alignment per triple can help in many ways. First, a triple can have several translations. Second, the system learned to generate related statements that might help rank the final statements. 

Finally, once we have all the translations to ConceptNet triples, we compute a score for each triple based on the frequency at which it appeared (several OpenIE triples generally generate the same closed triple) and the inverse rank among the predictions. More formally, we obtain the score of a triple $t$ using:
\begin{equation}
        \text{FinalScore}(t) = \sum_{\text{t' generates t}} \frac{\text{score}(t')}{\text{rank}(t', t) + 1}
\end{equation}
We will also consider two other scores in Section~\ref{sec:experiments}. The first only considers the open KB score part of the previous formula (a sum of scores), while the second only considers the ranks (a sum of reciprocal ranks). Here, it is essential to notice that the score of an open triple is provided by the open KB. Therefore, if the open KB is not good at scoring triples, we will inherit negative signals that we hope to compensate with the ranks.

In the end, we can generate a ranking for all our statements. Moreover, using a generative LM allows for having friendly properties missing in previous works. For example, it can adapt the subject and the object to match the new predicate. Besides, it can also correct the original statement if it contains a mistake (spelling or truth). Furthermore, it can inverse the subject and the object without additional help. Finally, it can generate multiple outputs from one input, bringing value to the end KB. We will demonstrate these properties in Section~\ref{sec:experiments}.



%% file: parts/experiment_setup.tex
\section{Experiment Setup}
\label{sec:experiment_setup}

\input{parts/evaluation}

\input{parts/baselines}

\subsection{Implementation}

We implemented the baselines using Python3 (except for AMIE, written in Java). For the generative LM, we used GPT-2-large given by Huggingface. We ran our code on machines with NVIDIA Quadro RTX 8000 GPUs. Finetuning a language model required a single machine for a maximum of two days. We used three training epochs in our experiments. However, mapping an open KB to ConceptNet was much longer and took up to 30 days on a single GPU. Nevertheless, the computations can easily be parallelized on several GPUs by splitting the input data, which allows us to speed up the process. 
In our experiments, we used Quasimodo and ASCENT++ as open KBs and mapped them to ConceptNet commonsense relations. We make the code and data available (\href{https://julienromero.fr/data/GenT}{julienromero.fr/data/GenT}).

%% file: parts/evaluation.tex
\subsection{Evaluation}
\label{sec:evaluation}

\subsubsection{Automatic Global Metrics}

To get a general understanding of the generated KB after the mapping, we compute the size of the KB. However, as the size is insufficient to evaluate the recall~\cite{quasimodo}, we consider that ConceptNet is our gold standard as humans filled it. Then, we measure the number of triples from ConceptNet we can generate. We call it the \textit{automatic recall}. Likewise, we create the \textit{automatic precision}:

\begin{equation}
        R_a(KB_{trans}) = \frac{|KB_{trans} \cap KB_{target}|}{|KB_{target}|} 
\end{equation}
\begin{equation}
        P_a(KB_{trans}) = \frac{|KB_{trans} \cap KB_{target}|}{|KB_{trans}|}
\end{equation}

As a part of the target KB can be used in the training dataset, we also define $\bar{R}_a$ as:

\begin{equation}
        \overline{R}_a(KB_{trans}) = \frac{|KB_{trans} \cap KB_{target} - D_{train}|}{|KB_{target} - D_{train}|} 
\end{equation}

We also define $\overline{P_a}$ following the same rationale as $\overline{R}_a$. However, this metric does not capture the ranking of our statements. The ranking is crucial in open KBs as these KBs are often noisy. Ideally, we want to have correct and important statements with a high score. We introduce metrics to measure that property. First, we will also use an \textit{automatic precision at $K$} where, instead of considering the entire $KB_{trans}$, we will only consider its top $K$ statements. However, these metrics do not consider the entire KB. Therefore, we introduce a generalized mean reciprocal rank of the final ranked KB as follows:

\begin{equation}
        MRR(KB_{trans}) = \frac{\sum_{KB_{trans}[i] \in KB_{target}} \frac{1}{i}}{\sum_{i \in [1, |KB_{target}|]} \frac{1}{i}}
\end{equation}

\begin{equation}
        \overline{MRR}(KB_{trans}) = \frac{\sum_{KB_{trans}[i] \in KB_{target}  - D_{train} } \frac{1}{i}}{{\sum_{i \in [1, |KB_{target} - D_{train}|]} \frac{1}{i}}}
\end{equation}

These metrics allow us to measure the recall, but it gives more weight to correct high-ranked statements.

All the metrics presented depend on the quality and coverage of the original open knowledge base. Therefore, when considering a translated KB, we prefer relative metrics where the metric is divided by the metric computed for the open knowledge base, ignoring the relations.

\subsubsection{Automatic Triple Alignment Metrics}

In Section~\ref{sec:alignment}, we suggested methods to align an open KB with a closed KB. These techniques generate a dataset of alignments that can be split into a training and a testing set used to evaluate the MRR, the precision (@K), and the recall (@K).

\subsubsection{Manual Metrics}

The automatic metrics we presented above are cheap to run but give a coarse approximation of the quality of the resulting knowledge. Therefore, we introduce manual metrics here. They are more expensive to run as they require human work but will provide a more precise evaluation.

\paragraph{Manual Triple Metrics} Inspired by~\cite{gashteovski2020aligning}, we would like to evaluate the quality of the triple mapping according to three parameters:

\begin{itemize}
    \item \textit{Correct mapping}: Is the generated triple a correct mapping of the open triple, i.e., is it semantically equivalent/related to the original triple?
    \item \textit{Correct prediction}: Is the resulting triple true? Independently of whether the mapping is correct, we would like to know if the resulting triple is accurate. This can be useful for several reasons. First, even if the mapping is incorrect, we would prefer that it does not hurt the quality of the knowledge base we construct next. Second, as the input triple may be noisy and incorrect, we would prefer that the system generates a correct statement rather than a correct mapping. Finally, if such a property holds, it will prove that the system has some cleaning properties that will help improve the quality of the open KB.
    \item \textit{Correct open triple}: Is the original open triple correct? This information will help evaluate what the system predicts depending on the quality of the input triple (see the point above). 
\end{itemize}

\paragraph{Knowledge Base Level Metrics}
Precision and recall are crude automated heuristics w.r.t.\ another data source. To evaluate the quality of novel CSK resources meaningfully, we rely on the \textit{typicality} notion of previous works~\cite{quasimodo,nguyen2021advanced}: We ask humans how often a statement holds for a given subject. Possible answers are: Invalid (the statement makes no sense) or Never / Rarely / Sometimes / Often / Always. Each answer has a score between 0 and 4 to compute a mean.





%% file: parts/baselines.tex
\subsection{Baselines}
\label{sec:baselines}

\subsubsection{Manual Mapping.}

For this baseline, we manually map the relations in an open KB to relations in ConceptNet. It is inspired by an idea from~\cite{nguyen2021refined}. Given a predicate \textit{p} in an open knowledge base, we ask humans to turn it into a predicate \textit{p'} in ConceptNet (including inverse relations). There are many relations in an open KB, so we only mapped the top relations. We also notice that, in many cases, a triples \textit{(s, p, o)} can directly be mapped to the triple \textit{(s, CapableOf, p + o)}. For example, \textit{(elephant, live in, Africa)} could be mapped to \textit{(elephant, CapableOf, live in Africa)}. If we cannot find a better translation, we default to this translation. This approach is a simple rule system. In our case, we annotated 100 predicates for Quasimodo and Ascent++. By doing so, we cover 82\% of triples in Quasimodo and 57\% of triples in Ascent++. 

\input{parts/rule_mining_old}

\subsubsection{Classification Task.}

For this baseline inspired by OpenKI~\cite{zhang2019openki}, we want to use a classifier to predict the ConceptNet relation of an open triple. Given a triple \textit{(s, p, o)} in an open KB, we want to predict a relation $p'$ (including inverse relations) in ConceptNet such that $(s, p', o)$ would be in ConceptNet. To do so, we used a classifier based on BERT~\cite{devlin2018bert} and trained it with a dataset created automatically (see Section~\ref{sec:alignment}). Building this dataset by hand would be possible, but it would take much time, and we would get problems getting enough examples for each predicate. Besides, we will use the same training dataset with the translation models.

%% file: parts/rule_mining_old.tex
\subsubsection{Rule Mining}

We propose a rule mining approach inspired by previous works~\cite{soderland2013open,soderland2010adapting,dutta2014semantifying}. Our method requires a training dataset of mappings. In our case, this dataset was constructed automatically (see Section~\ref{sec:alignment}) using the rule-based alignment method. The LM-based alignment is inappropriate as the subject and object must remain unchanged with the rule mining approach. Then, we construct a meta-knowledge base. Given a mapping from \textit{(s, p, o)} to \textit{(s', p', o')} represented by a unique identifier $M$, we generate the following statements:
\begin{itemize}
    \item \textit{(s' + M, p', o' + M)}. Here, we append the mapping identifier to the subject and object to prevent the rule mining system from using \textit{s'} and \textit{o'} as a constant.
    \item If $s'$ (resp. $o'$) matches $s$ ($s'$ is in $s$, after lemmatization and without stopwords), we create the statement \textit{(M, INSUBJ, s' + M)} (resp. \textit{(M, INSUBJ, o' + M)}).
    \item If $s'$ (resp. $o'$) matches $o$, we create the statement \textit{(M, INOBJ, s' + M)} (resp. \textit{(M, INOBJ, o' + M)}).
    \item For each hypernym $h$ of $s'$ (resp. $o'$) obtained with WordNet, we create the statement \textit{(s' + M, ISA, h)}. Here, we considered only hypernyms appearing at least ten times and in less than 50\% 
    \item We take the 100 more frequent tokens in the predicates of the open KB (e.g., ``in'', ``of'', ``be''). Then, if one of these tokens $t$ appears in $p$, we create the statement \textit{(M, CONTAINS, t)}.
\end{itemize}

Once we have this new KB, we use AMIE~\cite{lajus2020fast} to mine Horn rules of the form $B \Rightarrow r(x, y)$. The PCA confidence proposed in AMIE yields poor results. Therefore, we used the standard confidence. Besides, we modify the rule generation system so that the body cannot contain twice the same relation. These are the kind of rules we want in practice, and it allows us to mine the rules with many atoms much faster. Ultimately, we only keep rules with a confidence score greater than 0.5. An advantage of this method is that it provides high interpretability: For each final generation, we can see which open triples were used to generate it and which rules were applied.

\input{parts/tables/rule_mining}

We applied the rule mining system and obtained 72 rules for Quasimodo and 50 for ASCENT++. We show the best rules in Table~\ref{tab:rule_mining}. We observed that the system had difficulties generating good rules and finding suitable types for the subject or the object. This might come from several factors. First, the rules might not be complex enough and, therefore, cannot express a complicated mapping. Second, the standard confidence score might not be the best option. Indeed, if a rule applies very few times but is always right, it gets a high score, whereas it goes not give much information. Although we set a minimal support, we still observe this problem. Third, the complexity of the subjects and objects makes applying a taxonomy like WordNet difficult. So we will not get relevant type data. Finally, the system cannot adapt the subject and the object to match the new predicate better.

Looking at Table~\ref{tab:auto_rec_prec}, we observe that the system has trouble generalizing, i.e., generating statements not in the original training dataset. This is confirmed by the relatively low MRR, precision, and recall reported in Table~\ref{tab:auto_rec_prec_triple} when we evaluate the system on the testing set. The rules apply to a few cases, which explains the small size of the generated knowledge base.

%% file: parts/tables/rule_mining.tex
\begin{table}[!ht]
    \centering
    \begin{adjustbox}{width=\columnwidth,center}
    \begin{tabular}{|c|c|}
        \hline
        Rule & Confidence \\
        \hline
        \multicolumn{2}{|c|}{Quasimodo} \\
        \hline
        ?i  CONTAINS  cause  $\wedge$ ?i  INOBJ  ?b $\wedge$  ?i  INSUBJ  ?a $\wedge$  ?b  ISA  activity.n.01   $\Rightarrow$ ?a  Causes  ?b & 1.0 \\
        ?i  INOBJ  ?a $\wedge$  ?i  INSUBJ  ?b $\wedge$  ?b  ISA  structural\_member.n.01   $\Rightarrow$ ?a  DistinctFrom  ?b & 0.947580645 \\
        ?i  INOBJ  ?b $\wedge$  ?i  INSUBJ  ?a $\wedge$  ?b  ISA  representational\_process.n.01   $\Rightarrow$ ?a  HasA  ?b & 0.86440678 \\
        \hline
        \multicolumn{2}{|c|}{Ascent++} \\
        \hline
        ?i  INOBJ  ?a $\wedge$ ?i  INSUBJ  ?b $\wedge$ ?b  ISA  abstraction.n.01  $\Rightarrow$ ?a  Desires  ?b  & 0.769230769 \\
        ?i  INOBJ  ?a $\wedge$ ?i  INSUBJ  ?b $\wedge$ ?b  ISA  religious\_person.n.01  $\Rightarrow$ ?a  DistinctFrom  ?b   &     0.745454545 \\
        ?i  INOBJ  ?b $\wedge$ ?i  INSUBJ  ?a $\wedge$ ?b  ISA  administrative\_district.n.01 $\Rightarrow$ ?a  AtLocation  ?b &  0.737051793 \\

        \hline
        
    \end{tabular}
    \end{adjustbox}
    \caption{Top Rules Obtained With Rule Mining.}
    \label{tab:rule_mining}
    \dovspace
\end{table}

%% file: parts/results.tex
\section{Results and Discussion}
\label{sec:experiments}

This section will study several research questions investigating how our new model works. We will first look at the best mapping algorithm and then focus on finding the best alignment method, as this step of our pipeline has the most impact on the final result. Then, we will look at the properties of our model.




\input{parts/comparison_baselines}

\input{parts/tables/auto_rec_prec_rel}
\input{parts/tables/manual_annotation}

\input{parts/tables/auto_rec_prec_triples}

\input{parts/tables/so_conservation}

\input{parts/rq_mapping}

\input{parts/rq_alignment}
\input{parts/rq_properties}

\input{parts/direct_generation}

%% file: parts/comparison_baselines.tex
\subsection{Comparison With Baselines}

Table~\ref{tab:auto_rec_prec} shows the results of the automated metrics for all baselines. The first thing to notice is that the metrics seem ``low''. We recall that they are, in fact, relative to the open KB with the relations ignored, as mentioned in Section~\ref{sec:evaluation}. Therefore, they only have a relative interpretation. Even with the generous evaluation of the open KB, many metrics have a value of more than one, showing a significant improvement, particularly for the recall. For precision, a value less than one mainly comes with the growth of the KB size.

Our proposed approach clearly outperforms the various baselines. The basic models are not flexible and do not tackle the challenges we mentioned earlier. For manual mapping, the annotation process depends on humans and is not trivial, as translating a predicate often depends on the context. The classifier model performs better than the two other baselines when we look at the recall. Still, we observe problems to generalize as $\overline{R_a}$ is low.

%% file: parts/tables/auto_rec_prec_rel.tex
\begin{table*}[!ht]
    \centering

\begin{adjustbox}{width=\textwidth,center}
\begin{tabular}{|c|c|c|c|c|c|c|HHHc|c|c|c|c|}
        \textbf{KB}  & \textbf{Method} & \textbf{Training data} & $R_{a}$ & $\overline{R_{a}}$ & $P_{a}$ & $\overline{P_{a}}$ & $P_{a}@10$ & $P_{a}@100$ & $P_{a}@1000$ & $P_{a}@10k$ & $\overline{P_{a}}@10k$ & MRR & $\overline{MRR}$ & \textbf{Size} \\
        \hline \hline
        ConceptNet & KB itself & - & - & - & - & - & - & - & - & - & -  & - & - & 232,532 \\
        \hdashline
        Quasimodo & KB itself & - & 2.54\%$^*$ & - & 0.271\%$^*$ & - & 10\%$^*$ & 17\%$^*$ & 11\%$^*$ & 4.79\%$^*$ & -  & 8.32\%$^*$ & - & 5,930,628\\ 
        \hdashline
        Ascent++ & KB itself & - & 1.63\%$^*$ & - & 0.430\%$^*$ & - & 10\%$^*$ & 10\%$^*$ & 6.6\%$^*$ & 3.13\%$^*$ & - & 6.40\% $^*$  & - & 1,967,126\\ 
        \hline
        \hline
         \textbf{KB}  & \textbf{Method} & \textbf{Training data} & $R_{a, rel}$ & $\overline{R_{a, rel}}$ & $P_{a, rel}$ & $\overline{P_{a, rel}}$ & $P_{a, rel}@10$ & $P_{a, rel}@100$ & $P_{a, rel}@1000$ & $P_{a, rel}@10k$ & $\overline{P_{a, rel}}@10k$ & MRR$_{rel}$ & $\overline{MRR_{rel}}$ & \textbf{Size} \\
        \hline
        \hline
        \multirow{8}{*}{Quasimodo} & Manual Mapping~\cite{nguyen2021refined} & - & 0.231 & - & 0.103 & - & 1.000 & 0.471 & 0.455 & 0.315 & - & 0.592 & - & 4,925,792 \\
        & Rule Mining~\cite{soderland2013open,soderland2010adapting,dutta2014semantifying} & Rule-based & 0.161 & 0.006 & 0.509 & 0.020 & 3.000 & 1.235 & 0.645 & 0.365 & 0.004 & 1.259 & 0.002  & 689,146 \\
        & Classifier~\cite{zhang2019openki} & Rule-based & 0.752 & 0.042 & 0.299 & 0.016 & 2.000 & 1.000 & 0.855 & 0.672 & 0.002 & 1.419 & 0.001 & 5,478,028 \\
        \hdashline
         & GenT@1 & Rule-based & 1.465 & 0.425 & \textbf{0.771} & 0.217 & 8.000 & 4.471 & 4.127 & 3.361 & 0.201 & 4.816 & 0.098 & 4,135,349 \\
         & GenT@10 & Rule-based & 2.563 & 1.319 & 0.176 & 0.085 & 9.000 & 4.588 & 4.227 & \textbf{3.612} & 0.234 & \textbf{4.968} & 0.097 & 33,425,732 \\
         & GenT@10 & LM-based@10k & 2.370 & 1.677 & 0.347 & \textbf{0.235} & 2.000 & 2.294 & 3.227 & 2.777 & 0.357 & 2.505 & 0.069 & 15,647,853 \\
         & GenT@10 & LM-based@10k-INV & \textbf{2.787} & \textbf{1.933} & 0.241 & 0.162 & 1.000 & 1.059 & 1.391 & 1.939 & \textbf{0.660} & 1.333 & \textbf{0.216} & 25,798,594\\
         \hdashline
         & T5-GenT@10 & LM-based@10k-INV & 1.843 & 1.020 & 0.123 & 0.065 & - & - & - & 1.094 & 0.236 & 0.670 & 0.070 & 33,874,204 \\
        \hline
        \multirow{8}{*}{Ascent++} & Manual Mapping~\cite{nguyen2021refined} & - & 0.287 & - & 0.205 & - & 0.000 & 0.500 & 0.485 & 0.415 & - & 0.351 & - & 1,228,001 \\
         & Rule Mining~\cite{soderland2013open,soderland2010adapting,dutta2014semantifying} & Rule-based & 0.223 & 0.060 & 0.705 & 0.190 & 2.000 & 0.600 & 0.394 & 0.511 & 0.045 & 1.306 & 0.034 & 277,835\\
         & Classifier~\cite{zhang2019openki} & Rule-based & 0.663 & 0.180 & 0.340 & 0.105 & 2.000 & 1.200 & 0.652 & 0.649 & 0.026 & 0.784 & 0.016  & 1,722,441\\
        \hdashline
         & GenT@1 & Rule-based & 1.706 & 0.785 & \textbf{1.147} & \textbf{0.523} & 4.000 & 4.800 & 3.091 & 2.722 & 0.396 & 2.949 & 0.278  & 1,277,065\\
         & GenT@10 & Rule-based & 3.055 & 1.933 & 0.260 & 0.160 & 5.000 & 4.800 & 3.803 & 3.073 & 0.454 & 3.989 & 0.500  & 10,193,040\\
         & GenT@10 & LM-based@10k & 3.497 & 2.546 & 0.444 & 0.319 & 6.000 & 4.700 & 3.803 & \textbf{3.450} & 1.096 & \textbf{4.494} & 0.216  & 7,000,135 \\

         & GenT@10 & LM-based@10k-INV & \textbf{4.000} & \textbf{2.613} & 0.428 & 0.272 & 4.000 & 2.000 & 2.727 & \textbf{3.450} & \textbf{1.326} & 2.736 & \textbf{0.556}  & 8,305,861\\
        \hline
    \end{tabular}
\end{adjustbox}

    \caption{Automatic (Relative) Recall And Precision ($^*$ ignores the predicates).}
    \label{tab:auto_rec_prec}
\end{table*}

%% file: parts/tables/manual_annotation.tex
\begin{table}[ht]
    \centering

\begin{adjustbox}{width=0.6\columnwidth,center}
    \input{parts/tables/raw_typicality}
\end{adjustbox}

    \caption{Manual annotation.}
    \label{tab:manual_annotation}
    \dovspace
\end{table}

%% file: parts/tables/raw_typicality.tex
\begin{tabular}{|c|c|c|}
    \hline
        \textbf{KB} & \textbf{Alignment} & \textbf{Typicality} \\
    \hline
        ConceptNet & - & \textbf{3.18} \\
    \hline
        Quasimodo & - & 2.70 \\
        Quasimodo & Rule-based & \underline{2.91} \\
        Quasimodo & GenT@10k-INV & 2.88 \\
    \hline
        Ascent++ & - & 2.31 \\
        Ascent++ & Rule-based & 2.68 \\
        Ascent++ & GenT@10k-INV & \underline{2.88} \\
    \hline
    \end{tabular}

%% file: parts/tables/auto_rec_prec_triples.tex
\begin{table}[!ht]
    \centering

\begin{adjustbox}{width=\columnwidth,center}
\begin{tabular}{|c|c|c|c|c|c|c|c|c|c|}
        \hline
        \textbf{KB}  & \textbf{Method} & \textbf{Dataset} & \textbf{MRR} & \textbf{$R@1$} & \textbf{$R@5$} & \textbf{$R@10$} & \textbf{$P@1$} & \textbf{$P@5$} & \textbf{$P@10$}\\
        \hline

        \multirow{10}{*}{\rotatebox{90}{Quasimodo}} & Manual & Manual & 1.56$e^{-2}$ & 1.51\% & - & - & 1.56\% & - & - \\
         & Rule Mining & Rule-based & 5.96$e^{-2}$ & 5.55\% & 17.8\% & 24.8\% & 5.68\% & 3.63\% & 2.52\% \\
         & Classifier & Rule-based & 0.194 & 19.0\% & - & - & 19.4\% & - & - \\
         & GenT@10 & Rule-based & \textbf{0.381} & \textbf{31.6\%} & \textbf{46.7\%} & \textbf{49.5\%} & \textbf{31.1\%} & \textbf{9.67\%} & \textbf{5.12\%}  \\
         & GenT@10 & LM-based@10k & 0.279 & 23.1\% & 34.5\% & 36.9\% & 23.1\% & 6.91\% & 3.69\% \\
         & GenT@10 & LM-based@100k & 0.319 & 27.5\% & 38.0\% & 39.8\% & 27.5\% & 7.60\% & 3.98\% \\
         & GenT@10 & LM-based@1k-INV & 0.211 & 15.4\% & 26.9\% & 34.6\% & 15.4\% & 5.38\% & 3.46\% \\
         & GenT@10 & LM-based@10k-INV & 0.123 & 8.48\% & 17.1\% & 20.4\% & 8.77\% & 3.51\% & 2.09\% \\
         & T5-GenT@10 & LM-based@10k-INV & 0.129 & 10.0\% & 16.6\% & 19.9\% & 10.1\% & 3.40\% & 2.05\% \\
        \hline
    \end{tabular}
\end{adjustbox}

    \caption{Automatic Triple Alignment MRR, Recall And Precision (as usually defined).}
    \label{tab:auto_rec_prec_triple}
    \dovspace
\end{table}

%% file: parts/tables/so_conservation.tex
\begin{table}[!ht]
    \centering
    \begin{adjustbox}{width=\columnwidth,center}
    \begin{tabular}{|c|c|c|c|c|c|c|c|c|c|}
        \hline
         & \multicolumn{3}{|c|}{\textbf{First gen.}} & \multicolumn{3}{|c|}{\textbf{At least one gen.}} & \multicolumn{3}{|c|}{\textbf{All gens.}} \\
         \hline
        \textbf{Alignment} & \textbf{S} & \textbf{O} & \textbf{SO} & \textbf{S} & \textbf{O} & \textbf{SO} & \textbf{S} & \textbf{O} & \textbf{SO} \\
        \hline
        Rule-based & 36.8\% & 48.5\% & \textbf{26.6\%} & \textbf{57.9\%} & 76.2\% & \textbf{48.3\%} & 25.0\% & \textbf{35.3\%} & \textbf{12.4\%} \\
        LM-based@1k & 27.5\% & 21.0\% & 7.37\% & 45.3\% & 41.7\% & 17.9\% & 22.3\% & 12.3\% & 2.37\% \\
        LM-based@1k-INV & \textbf{38.7\%} & \textbf{53.6\%} & 22.4\% & 55.2\% & \textbf{77.5\%} & 42.0\% & \textbf{27.0\%} & 31.9\% & 5.84\% \\
        \hline
    \end{tabular}
    \end{adjustbox}
    \caption{SO Conservation For Quasimodo.}
    \label{tab:so_conservation}
    \dovspace
\end{table}

%% file: parts/rq_mapping.tex
\subsection{What is the best mapping method?}

In Table~\ref{tab:auto_rec_prec}, we present the main results of our paper with a comparison with other baselines. We called our approach \textbf{GenT@K} (for Generative Translation with K as a parameter for the number of closed triples per open triple), and we show here the results from some of the best automatic alignments we found (more about this later). GenT@K outperforms the other baselines for both Quasimodo and Ascent++. For the rule mining approach, $P_a$ has a high value. This is due to the generated KB's small size, which comes from the difficulty of finding good rules. However, when we look at $\overline{P_a}$, we see that the rule mining approach clearly does not generalize.

More generally, GenT methods really shine when we look at the metrics that do not consider training data. It confirms our hypothesis that we can build models that generalize better and can adapt to more situations with generative translation. 

\subsubsection{What is the influence of the number of generations?}

\begin{figure}
    \centering
    \includegraphics[width=0.9\columnwidth]{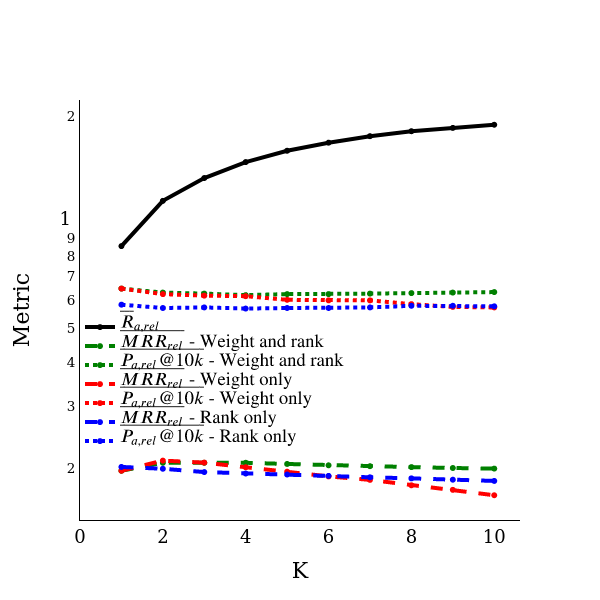}
    \caption{Impact of the number of generated triples K for each open triple - Quasimodo - LM-based@10-INV}
    \label{fig:impact_k_quasimodo}
\end{figure}

A key parameter for the GenT@K method is the number of generations $K$ we consider for each triple. In Figure~\ref{fig:impact_k_quasimodo}, we make $K$ vary from 1 to 10. We see that $\overline{R_a}$ continuously increases, and this was expected: The more generations, the higher the chance to overlap with ConceptNet. However, this metric has not plateaued yet, indicating that we could increase the number of generations (but with a higher computational cost).

A good recall is not helpful if we cannot differentiate good triples from bad triples. So, it is essential to also look at the precision. Here, we observe that the $\overline{MRR}$ and the $\overline{P_a@10k}$ remain stable when considering a score with the weight and the rank. On the one hand, it is a good sign because it shows that we do not add noise, but, on the other hand, we would have wished that the precision metrics increase thanks to the corroboration. This result suggests we must design a more advanced scoring method to leverage the multiple generations fully.

Looking more closely at our scoring methods, we can see that using both the weight and the rank gives better results than using them separately. This suggests that they are both critical elements of the scoring function.

%% file: parts/rq_alignment.tex
\subsection{What is the best alignment method?}


In Section~\ref{sec:alignment}, we presented two automatic alignment methods. The first is based on a rule system, whereas the second aligns with the closest triples in a latent space using embeddings. We refer to the first as Rule-based and the second as LM-based@K(-INV) when we used top K statements of the complete dataset obtained by aligning each open triple with a close triple (INV means we align each close triple with an open triple).


\input{parts/tables/manual_alignment_evaluation}

\subsubsection{Do they allow the model to generate accurate alignments?}

Table~\ref{tab:auto_rec_prec_triple} gives the performance of the model on a test dataset derived from the complete dataset. Therefore, it is not the same for all models and depends on the alignment method. Still, it gives us some valuable insights. We can see that the Rule-based alignment is the easiest to learn. This is due to the strong correlation created by the rules between the open and the closed triples. According to the metrics, the INV methods perform worse than non-INV ones. A reason might be that the INV alignment has more diversity: A triple from ConceptNet can appear only once in the dataset (we align each close triple with a single open triple). Therefore, it might be harder to learn.

Table~\ref{tab:so_conservation} shows the conservation of the subject S and object O during the generation phase. We want to observe if they remain the same for the first generation, for at least one generation, or for all generations. The rule-based system encodes these constraints and should therefore outperform the other baselines. However, interestingly, we observe that the INV methods have excellent conservation, competing with the rule-based system (except for SO conservation), and largely beating non-INV alignments. This is surprising as it contains no prior constraint. It is a property that we expect from a good alignment method as we do not want the generated close triples to diverge from the original triples.

All these evaluations are automatic and only approximate the model's capabilities. We additionally performed manual annotations of the generations to check if the generated close triples are correct alignments (according to semantic relatedness, as discussed in Section~\ref{sec:evaluation}). We sampled 300 triples from the top 10k triples in Quasimodo and looked at the first generation for three models. The results are presented in Table~\ref{tab:manual_alignment_evaluation}. We observed that the rule-based and INV alignments have similar performances for generating related close triples. Only the non-INV model underperforms, which matches what we noticed for SO conservation. Here, semantic relatedness is relatively low because it is quite constraining. However, we observe that the generated triples share most of the time part of the subject or object with the original triple.

\subsubsection{What is the impact of the training dataset size?}

\begin{figure}
    \centering
    \includegraphics[width=0.7\columnwidth]{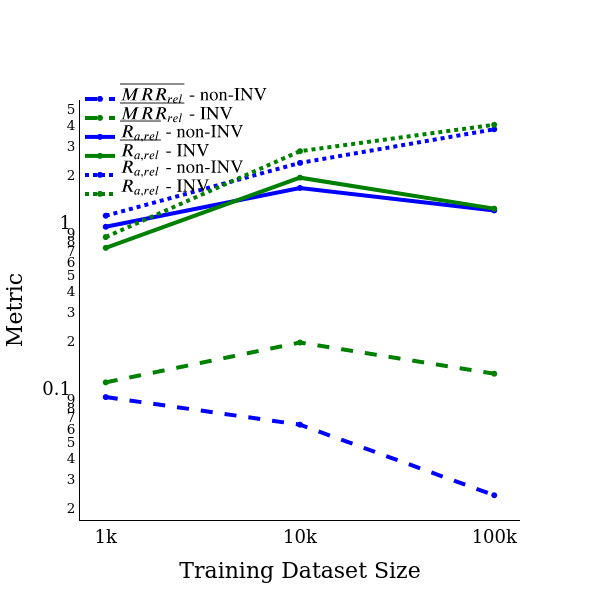}
    \caption{Impact of Training Dataset Size - Quasimodo}
    \label{fig:impact_dataset_size_quasimodo}
\end{figure}

We sampled the top-$K$ samples with $K \in \{1k,10k,100k\}$ in the training dataset (see Figure~\ref{fig:impact_dataset_size_quasimodo}) and picked the best size. We observed that the model performs best for 10k samples. Note that finding the optimal size would take too much time as, with these metrics, we need to generate the entire mapping. The testing dataset used in Table~\ref{tab:auto_rec_prec_triple} gives a faster heuristic for finding a good K.

%% file: parts/tables/manual_alignment_evaluation.tex
\begin{table}[!ht]
    \centering

\begin{adjustbox}{width=\columnwidth,center}
    \input{parts/tables/raw_manual_alignment}
\end{adjustbox}

    \caption{Manual Alignment Evaluation On Quasimodo.} 
    \label{tab:manual_alignment_evaluation}
    \dovspace
\end{table}

%% file: parts/tables/raw_manual_alignment.tex
\begin{tabular}{|c|c|c|c|c|}
    \hline
        \textbf{Dataset} & \textbf{Sem. Rel.} & \textbf{Open Correct} & \textbf{Close Correct} & \textbf{Both Correct} \\
    \hline
        Rule-based & 53.0\% & 85.3\% & 69.3\% & 64.8\% \\
        GenT@10k & 45.7\% & 85.3\% & 75.7\% & 68.0\% \\
        GenT@10k-INV & \textbf{55.3\%} & 85.3\% & \textbf{77.3\%} & \textbf{69.7\%} \\
    \hline
    \end{tabular}

%% file: parts/rq_properties.tex
\subsection{What are the properties of our model?}

In Section~\ref{sec:introduction} and Section~\ref{sec:problem}, we described properties we want our new system to have, such as high recall and precision, flexibility thanks to the LM, and good generalization. We will investigate these advantages here in more detail.

\input{parts/rq_advantages}
\input{parts/rq_quality}
\input{parts/rq_generalize_other}
\input{parts/rq_sentences}

%% file: parts/rq_advantages.tex
\subsubsection{What are the advantages of using a generative translation model?}

When we use traditional models like manual mapping, rule mining, or classifiers, we encounter problems coming from their lack of flexibility: The subject and object often remain unchanged, the mapping is often unique, mistakes are not corrected, etc. With a generative translation model, we can tackle some of these challenges. In Table~\ref{tab:examples}, we present examples of nice and unique properties we observed in the final results. This shows that using LMs with open KBs lets us get the best of both worlds. LMs are more flexible and contain knowledge that is not easily extractable~\cite{petroni2019}. The open KB helps guide the LM.

\input{parts/tables/examples}

%% file: parts/tables/examples.tex
\begin{table*}[ht]
    \centering
\begin{adjustbox}{width=0.7\textwidth,center}
\begin{tabular}{|l|c|c|}
        \hline
        \textbf{Property} & \textbf{Open triple} & \textbf{Generated closed triple} \\
        \hline
        The mapping depends on & & \\
        the context & (elephant, has\_property, decorated) & (elephant, ReceivesAction, decorated) \\
        \hline
        Adapt the subject/object & (elephant, be in, africa killed) & (elephant, AtLocation, africa) \\
         & (doctor, write, prescription) & (doctor, CapableOf, prescribe) \\
         & (doctor, be in, training) & (doctor, AtLocation, medical school) \\
         & (doctor, keep, from getting sick) & (doctor, CapableOf, keep patient healthy) \\
        \hline
        Inversion subject/object & (elephant, have, tusks) & (tusk, PartOf, elephant) \\
        \hline
        Correct the subject/object & (elephant, have, ivory tusks answers) & (elephant, HasA, ivory tusk) \\
        \hline
        Put the predicate in the object & (doctor, wear, coat) & (doctor, CapableOf, wear coat) \\
        \hline
        Good triples from nonsense & (doctor, has\_property, as a career) & (doctor, HasProperty, well respected) \\
        \hline
        Generate related triples & (doctor, wear, mask) & (doctor, CapableOf, wear lab coat) \\
        \hline
        Turn s/o \ into standard form & (apartment, be in, nyc) & (apartment, AtLocation, new york city)\\
        \hline
    \end{tabular}
\end{adjustbox}

    \caption{Examples of Mappings from GenT.}
    \label{tab:examples}
\end{table*}

%% file: parts/rq_quality.tex
\subsubsection{Can we improve the quality of an open KB with a generative translation model?}

To evaluate the evolution of the quality of an open KB, we asked humans to annotate the typicality of statements. We sampled 300 statements out of the top 10k statements for each KB and then computed the mean typicality. The results are reported in Table~\ref{tab:manual_annotation}. As we can see, the generative translation methods significantly improve the quality of the statements. The best-performing alignment method seems to depend on the open KB. As expected, ConceptNet still outperforms our approach as it was manually generated. However, it does not have the same scaling capabilities.

%% file: parts/rq_generalize_other.tex
\subsubsection{Can GenT generalize across open KBs?}

Table~\ref{tab:comp_other_openkb} shows the results of models trained on one open KB, Quasimodo or Ascent++, and used to generate a closed KB from triples in another open KB. We chose the LM-based@10k-INV alignment. In most cases, the original model trained with the same open KB outperforms the foreign model. This is understandable as the data sources and processing steps used to generate the open KBs differ, and therefore the style of the open triples is different. So, the model might have difficulties adapting. Still, the new results are close to the original ones, showing that we can have the reusability of our models with entirely new data. Finally, some metrics seem less impacted by the change of the original open KBs. From what we can see, the ranking capabilities, expressed through $P_a$ and $MRR$, vary but not necessary for the worst. It shows that the generation and the scoring stage allow selecting good close triples, whatever the new data is.

\input{parts/tables/comparison_other_openkb_model}

%% file: parts/tables/comparison_other_openkb_model.tex
\begin{table*}[!htbp]
    \centering

\begin{adjustbox}{width=0.6\textwidth,center}
\begin{tabular}{|c|c|c|c|c|cHHH|c|c|c|c|}
     KB  & Method & $R_{a, rel}$ & $\overline{R}_{a, rel}$ & $P_{a, rel}$ & $\overline{P_{a, rel}}$ & $P_{a, rel}@10$ & $P_{a, rel}@100$ & $P_{a, rel}@1000$ & $P_{a, rel}@10k$ & $\overline{P_{a, rel}@10k}$ & $MRR_{rel}$ & $\overline{MRR_{rel}}$\\
        \hline
        \textcolor{gray}{Quasimodo} & \textcolor{gray}{GenT@1} & \textcolor{gray}{1.54} & \textcolor{gray}{0.89} & \textcolor{gray}{1.15} & \textcolor{gray}{0.65} & \textcolor{gray}{20\%} & \textcolor{gray}{17\%} & \textcolor{gray}{15.3\%} & \textcolor{gray}{1.95} & \textcolor{gray}{0.65} & \textcolor{gray}{1.27} & \textcolor{gray}{0.21} \\
        \textcolor{gray}{Quasimodo} & \textcolor{gray}{GenT@10} & \textcolor{gray}{2.79} & \textcolor{gray}{1.93} & \textcolor{gray}{0.24} & \textcolor{gray}{0.16} & \textcolor{gray}{10\%} & \textcolor{gray}{18\%} & \textcolor{gray}{15.3\%} & \textcolor{gray}{1.94} & \textcolor{gray}{0.66} & \textcolor{gray}{1.33} & \textcolor{gray}{0.22}\\
        \hline
        Quasimodo & GenT@1 & 1.35 & 0.81 & 0.77 & 0.46 & 70\% & 67\% & 37.3\% & 0.73 & 0.76 & 4.28 & 0.98 \\
        Quasimodo & GenT@10 & 2.44 & 1.77 & 0.18 & 0.13 & 70\% & 71\% & 38.6\% & 2.92 & 0.84 & 4.41 & 0.96 \\
        \hline
        \textcolor{gray}{Ascent++} & \textcolor{gray}{GenT@1}& \textcolor{gray}{2.06} & \textcolor{gray}{1.09} & \textcolor{gray}{1.84} & \textcolor{gray}{0.96} & \textcolor{gray}{30\%} & \textcolor{gray}{23\%} & \textcolor{gray}{16.8\%} & \textcolor{gray}{3.32} & \textcolor{gray}{1.23} & \textcolor{gray}{2.80}  & \textcolor{gray}{2.61$e^{-6}$} \\
        \textcolor{gray}{Ascent++} & \textcolor{gray}{GenT@10} & \textcolor{gray}{4.00} & \textcolor{gray}{2.61} & \textcolor{gray}{0.43} & \textcolor{gray}{0.27} & \textcolor{gray}{40\%} & \textcolor{gray}{20\%} & \textcolor{gray}{18\%} & \textcolor{gray}{3.45} & \textcolor{gray}{1.33} & \textcolor{gray}{2.74}  & \textcolor{gray}{0.56} \\
        \hline
        Ascent++ & GenT@1 & 1.95 & 0.77  & 1.28  & 0.50 & 80\%  & 56\%  & 25.2\%  & 3.17 & 0.38 & 4.69 & 0.08 \\
        Ascent++ & GenT@10 & 3.62 & 1.96 & 0.30 & 0.16 & 80\%  & 65\%  & 32.4\%  & 3.87 &  0.44 & 5.37 & 0.10 \\
        \hline
    \end{tabular}
\end{adjustbox}

    \caption{Performances when evaluating with a model trained for another KB. (grey = the original results)}
    \label{tab:comp_other_openkb}
\end{table*}

%% file: parts/rq_sentences.tex
\subsubsection{Generalization To Sentences}

In Table~\ref{tab:sentences}, we took sentences or paragraphs from several sources (Wikipedia, New York Times, GenericsKB~\cite{bhakthavatsalam2020genericskb}) and used our model trained on Quasimodo with the LM-based@10k-INV alignment method. Surprisingly, the model can correctly extract knowledge from sentences. This could lead to several interesting future works: Information extraction directly from sentences, aligning sentences rather than open triples, or commonsense inference.

\input{parts/tables/sentences}

%% file: parts/tables/sentences.tex
\begin{table*}[!htbp]
    \centering
\begin{adjustbox}{width=0.6\textwidth,center}
\begin{tabular}{|l|c|}
        \hline
        \textbf{Source} & \textbf{First Generation} \\
        \hline
        Elephants are the largest existing land animals. & (elephants, DefinedAs, largest land animal) \\
                \hline

        A lawyer or attorney is a person who practices law. & (lawyer, CapableOf, represent client) \\
                \hline

        Elon Musk Races to Secure Financing for Twitter Bid. & (elon musk, CapableOf, bid for twitter) \\
                \hline

        South Africa’s Government Shifts to Rebuilding After & (people, CapableOf, die from flooding) \\
        Disastrous Flooding. Nearly 4,000 homes have been  & \\
        destroyed and more than twice as many damaged in the  & \\
        Durban area after a week of punishing rains and mudslides.  & \\
        The death toll is now 448, with about four dozen people & \\
        unaccounted for.& \\
                \hline

        Some air pollutants fall to earth in the form of acid rain. & (air pollution, CapableOf, cause acid rain) \\
        \hline
    \end{tabular}
\end{adjustbox}

    \caption{Examples of Generations From Sentences.}
    \label{tab:sentences}
\end{table*}

%% file: parts/direct_generation.tex
\subsection{How does GenT compare with direct LM generation methods?}

As previous works like LAMA~\cite{petroni2019} suggested, a powerful language model could serve as a knowledge base. Then, aligning this ``knowledge base'' with a target knowledge base requires finetuning the language model. COMET~\cite{comet-atomic-2020} finetunes GPT-2~\cite{radford2019language} to generate triples in ConceptNet. Here, we consider two kinds of input: A subject alone (denoted as COMET S) or a subject/predicate pair (designated as COMET SP). COMET initially accepted only subject/predicates pairs. However, it makes the generation of relevant triples harder as it is not always possible to associate all subjects to all predicates (for example, ``elephant'' and ``HasSubEvent''). Then, we generate ten candidate statements for each subject or subject/predicate pair in ConceptNet. They all come with a generation score that we use for an overall ranking. In addition to the raw COMET, we used the translation models described above to generate a KB (GenT COMET). The inputs are the same as COMET. We additionally parse the output to keep the triple on the right of the \textit{[SEP]} token.  

It turns out that our translation model is a clever scheme in between traditional IE-based KB construction and a general COMET-style generation. It overcomes the limitation of IE that requires a text as input (it can generate more triples without requiring that each is seen in input text). It also tackles some COMET challenges by providing more robust guidance on what to generate based on the input triples.

In Table~\ref{tab:direct_generation}, we observe that GenT consistently outperforms COMET in all metrics but $R_a$. This is easily understandable: As COMET does not require alignments, we can get 5 to 10 times more training data than the translation models. These data points are guaranteed to represent different ConceptNet triples (not necessarily the case for the translation models). However, if we look at $\bar{R}_a$, the translation models generalize better. Besides, we have a ranking capability lacking in the original COMET. This could be explained by the fact that the translation models first try to generate an open triple closer to natural language and then map this triple to ConceptNet. Therefore, it can better leverage its prior knowledge to focus on what is essential.

\input{parts/tables/comet}

%% file: parts/tables/comet.tex
\begin{table*}[!htbp]
    \centering

\begin{adjustbox}{width=0.8\textwidth,center}
\begin{tabular}{|c|c|c|c|c|c|c|HHHc|c|c|c|}
        KB  & Method & Dataset & $R_a$ & $\overline{R}_a$ & $P_a$ & $\overline{P_a}$ & $P_a@10$ & $P_a@100$ & $P_a@1k$ & $P_a@10k$ & $\overline{P_a@10k}$ & MRR & $\overline{MRR}$\\
        \hline
        Quasimodo & GenT COMET S & Rule-based & 1.09\% & 0.222\% & \textbf{3.72\%} & 0.137\% & \textbf{100\%} & \textbf{83\%} & 42.2\% & \textbf{13.3\%} & 1.66\% &\textbf{46.6\%} & 0.46\% \\ 
        Quasimodo & GenT COMET SP & Rule-based & 2.33\% & 0.803\% & 0.403\% & 0.782\% & 30\% & 24\% & 18.6\% & 11.3\% & 1.24\% & 14.7\% & 0.41\% \\ 
        
        Quasimodo & GenT COMET S & LM-based@10-INV & 0.657\% & 0.285\% & 2.20\% & 0.975\% & 20\% & 17\% & 15.8\% & 7.14\% & 2.86\% & 17.2\% & 8.57\% \\ 
        Quasimodo & GenT COMET SP & LM-based@10-INV & 1.71\% & 1.07\% & 0.261\% & 0.163\% & 20\% & 21\% & 15.5\% & 6.46\% & 3.12\% & 12.9\% & 0.60\% \\ 
        \hline
        Ascent++ & GenT COMET S & Rule-based & 0.977\% & 0.358\% & 3.10\% & \textbf{1.15\%} & \textbf{100\%} & 81\% & \textbf{43\%} & 12.5\% & 3.90\% & 47.3\% & 4.45\% \\ 
        Ascent++ & GenT COMET SP & Rule-based & 2.15\% & \textbf{1.11\%} & 0.345\% & 0.177\% & 100\% & 74\% & 38.7\% & 12.6\% & 4.14\% & 20.0\% & \textbf{11.2\%}\\ 
        
        Ascent++ & GenT COMET S & LM-based@10-INV & 0.825\% & 0.326\% & 2.74\% & 0.326\% & 20\% & 26\% & 28.8\% & 8.94\% & 3.50\% & 11.6\% & 2.09\% \\ 
        Ascent++ & GenT COMET SP & LM-based@10-INV & 2.09\% & 1.10\% & 0.340\% & 0.178\% & 40\% & 24\% & 19.6\% & 10.7\% & \textbf{4.69\%} & 15.7\% & 5.84\%\\ 
        \hline
        - & Comet S & ConceptNet & 1.11\% & 0.144\% & 2.96\% & 0.401\% & 20\% & 25\% & 20.1\% & 7.65\% & 0.891\% & 11.3\% & 0.12\% \\ 
        - & Comet SP & ConceptNet & \textbf{2.87\%} & 0.504\% & 0.179\% & 0.504\% & 10\% & 2\% & 3.2\% & 3.36\% & 0.510 & 2.57\% & 0.05\% \\ 
        \hline
    \end{tabular}
\end{adjustbox}

    \caption{Direct Generation Comparison, non-relative metrics}
    \label{tab:direct_generation}
\end{table*}

%% file: parts/conclusion.tex
\section{Conclusion}

We studied the problem of mapping an open commonsense knowledge base to a fixed schema. We proposed a generative translation approach that carries novel properties such as flexibility and cleaning ability. In the process, we compared different ways to create training data and analyzed their advantages and disadvantages. Finally, we experimentally verified the strengths of the proposed approach both in automated and manual evaluation.

We provided the first solution for the mapping task, and there is still room for improvement. For example, we could study how to adapt state-of-the-art translation models. We could also check how the output of the generative model can be constrained to provide closed triples that are not too far from the original triples. Also, as we observed that LM-based models have cleaning capabilities, we could include a negative sample in the training dataset to predict cases where a triple has no translation (e.g. because it is incorrect).

We provide mappings to ConceptNet of Quasimodo and Ascent++ as additional resources in addition to the code and input data (\href{https://julienromero.fr/data/GenT}{julienromero.fr/data/GenT}). We hope they will help improve tasks such as commonsense question answering that currently use ConceptNet, which can sometimes be problematic as some of these datasets are constructed from ConceptNet (e.g., CommonsenseQA~\cite{talmor2018commonsenseqa}).